\documentclass[journal]{IEEEtran}
\usepackage{amsmath,amsfonts}
\usepackage{algorithmic}
\usepackage{algorithm}
\usepackage{array}
\usepackage[caption=false,font=normalsize,labelfont=sf,textfont=sf]{subfig}
\usepackage{textcomp}
\usepackage{stfloats}
\usepackage{url}
\usepackage{verbatim}
\usepackage{graphicx}
\usepackage{cite}
\usepackage{multirow}
\usepackage[utf8]{inputenc} 
\usepackage[T1]{fontenc}    
\usepackage{url}            
\usepackage{booktabs}       
\usepackage{amsfonts}       
\usepackage{nicefrac}       
\usepackage{microtype}      
\usepackage{xcolor}         
\usepackage{times}
\usepackage{soul}
\usepackage{url}
\usepackage[hidelinks]{hyperref}
\usepackage[small]{caption}
\usepackage{graphicx}
\usepackage{amsmath}
\usepackage{amsthm}
\usepackage{booktabs}
\usepackage{algorithm}
\usepackage{algorithmic}
\usepackage[switch]{lineno}
\usepackage{enumitem}
\usepackage{adjustbox}
\usepackage[framemethod=TikZ]{mdframed}

\usepackage{calc}

\newtheorem{definition}{Definition}
\newtheorem{proposition}{Proposition}

\mdfsetup{
    skipabove=1em,
    skipbelow=1em,
    linecolor=black,
    linewidth=1pt,
    innertopmargin=6pt,
    innerbottommargin=6pt
}
\begin{document}
\title{Architectural Scaling Surpass Basis Complexity? Efficient KANs with Single-Parameter Design}

\author{Zhijie Chen, Xinglin Zhang, Hongshu Guo, Yue-Jiao Gong
\thanks{The authors are with the School of Computer Science and Engineering, South China University of Technology (e-mail: zhijiechencs@gmail.com; zhxlinse@gmail.com; guohongshu369@gmail.com; gongyuejiao@gmail.com).}%
}

\maketitle

\begin{abstract}
	The landscape of Kolmogorov-Arnold Networks (KANs) is rapidly expanding, yet lacks a unified theoretical framework and a clear principle for efficient architecture design. This paper addresses these gaps with three core contributions. First, we introduce the Universal KAN (Uni-KAN) framework, a novel abstraction that formally unifies all KAN-style networks through dense and sparse representations. We prove their interchangeability and provide an open-source library for this framework, facilitating future research. Second, we propose the Efficient KAN Expansion (EKE) Hypothesis, a design philosophy positing that allocating parameters to architectural scaling rather than basis function complexity yields superior performance. Third, we present Single-Parameter KANs (SKANs), a family of ultra-lightweight networks that embody the EKE Hypothesis. Our comprehensive experiments provide the first strong empirical validation for the theoretical necessity of basis function smoothness for stable training. Furthermore, SKANs demonstrate state-of-the-art performance, improving F1 scores by up to 6.51\% and reducing test loss by 93.1\%, while achieving up to 6x faster training speeds compared to existing KAN variants. These results establish a robust framework, a guiding hypothesis, and a practical methodology for designing the next generation of efficient and powerful neural networks. The code is accessible at \url{https://anonymous.4open.science/r/SKAN-EBBB/}.
\end{abstract}


\section{Introduction}

The pursuit of efficient nonlinear representations forms the core challenge of modern neural architecture design. 
While Multilayer Perceptrons (MLPs) dominate practical applications, their fixed activation patterns and opaque parameter interactions limit both efficiency and interpretability.
Kolmogorov-Arnold Networks (KANs)~\cite{liu_kan_2024} have emerged as a promising alternative. Drawing inspiration from the Kolmogorov-Arnold representation theorem, KANs replace fixed activation functions with learnable nonlinear basis functions, which are then combined through summation. 
This architectural design enables independent visualization and symbolic representation of each nonlinear unit, establishing a foundation for enhanced model interpretability. 

The advent of KANs has sparked significant research enthusiasm, driving swift diversification in architectural design~\cite{somvanshi2024surveykolmogorovarnoldnetwork}. Initially anchored to spline-based basis functions (termed Spl-KAN in this work), the KAN paradigm has since evolved into a number of notable variants. For instance: 
Wav-KAN \cite{bozorgasl_wav-kan_2024} incorporates wavelet function, Fast-KAN \cite{li_kolmogorov-arnold_2024} employs radial basis function, and FourierKAN \cite{FourierKAN} adopts fourier transform. 
\begin{figure*}[t]
	\centering
	\includegraphics[width=\textwidth]{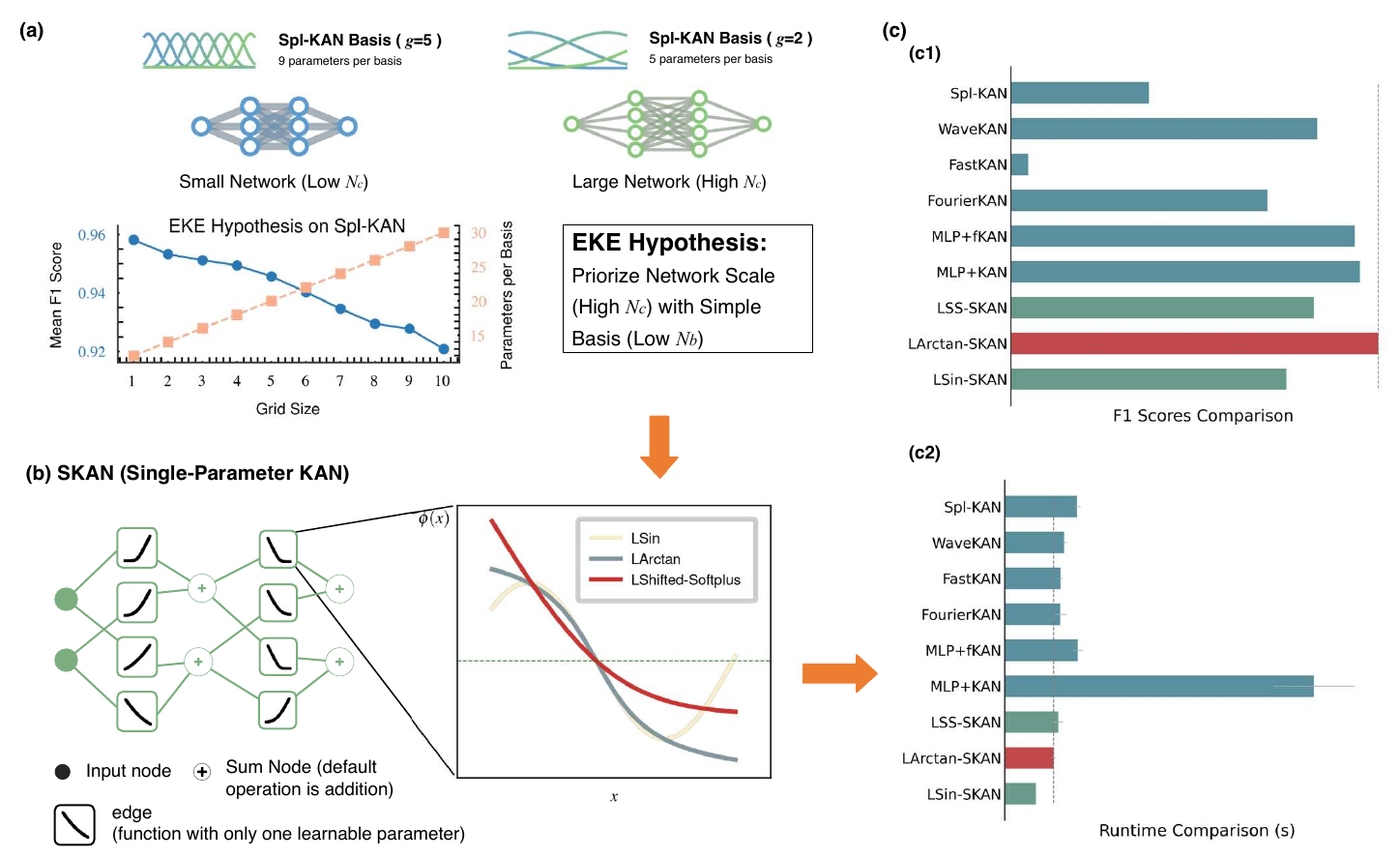}
	\caption{Visualization of SKANs' architecture and its key components. \textbf{Top left (a):} Basis function from Spl-KAN, debate between simple and complex basis and the preliminary experimental results demonstrating the EKE hypothesis. \textbf{Bottom left (b):} Visualization of a [2,2,2] SKAN network and bases of SKANs. \textbf{Right (c):} Part of SKANs' excellent performance compared to other models.}
	\label{fig:overview}
\end{figure*}
However, this rapid expansion has led to two fundamental challenges: a lack of a common language to describe and implement these varied architectures, and a design ethos that often prioritizes increasing the complexity of individual basis functions. This focus on basis complexity can lead to suboptimal parameter allocation and hinder network scalability, a critical factor for tackling complex problems. For example, the top left of \figurename~\ref{fig:overview} demonstrates that larger grid sizes in Spl-KAN increase parameter counts per basis but deliver worse performance when keeping total parameters the same. 

In this work, we address these challenges by starting from unification to efficiency. We first introduce the \textbf{Universal KAN (Uni-KAN)} framework, a formal abstraction capable of representing most KAN-style networks. This framework provides a theoretical and computational foundation for systematically designing and analyzing KANs. Building on this, we introduce a new design philosophy, which we term the \textbf{Efficient KAN Expansion (EKE) Hypothesis}:
``\textit{For a fixed parameter budget, prioritizing resource allocation toward network topology scaling (depth/width) over basis function complexity induces dominance in functional expressivity and parameter efficiency.}'' 

Guided by this hypothesis, we develop \textbf{Single-Parameter KANs (SKANs)}, a family of models that utilize ultra-lightweight basis functions with only a single learnable parameter. This minimalist approach maximizes the parameter budget available for architectural expansion. Our work provides comprehensive experimental validation for our claims, demonstrating the power of this new paradigm across multiple domains.

We summarize our contributions as follows:
\begin{enumerate}
	\item \textbf{The Uni-KAN Framework:} We propose a novel, unified framework that formally defines and standardizes KAN-like architectures. We prove the equivalence between its flexible (sparse) and computationally efficient (dense) representations and provide a corresponding open-source library to accelerate future research.
	
	\item \textbf{The EKE Hypothesis:} We introduce the EKE hypothesis for developing KAN variants, which advocates allocating parameters toward network scale expansion rather than increasing basis function complexity. This hypothesis diverges from prevailing methodologies, offering new insights to the scalability-performance trade-off of KANs. We provide extensive empirical evidence suggesting that architectural scaling with simpler functions is a more parameter-efficient path to high performance.
	
	\item \textbf{SKANs with Ultra-Lightweight Bases:} Based on EKE, we suggest a SKAN family with different basis functions. Specifically, three parameter-efficient basis functions are innovated: a) the LSS function for general-purpose applications, 2) the LArctan function optimized for classification, and 3) the LSin function for efficient computation. Each function contains only one learnable parameter, enabling efficient network expansion while maintaining interpretability.
	
	\item \textbf{Comprehensive Experimental Validation:} Experimental results reveal SKANs' superior performance across multiple domains: On the MNIST dataset, LArctan-SKAN improves F1 scores by 6.51\% and achieves over 6x faster training compared to mainstream KAN variants. For differential equation solving, LSS-SKAN-ODE reduces test loss by 93.1\% and standard deviation by 98.4\% versus Neural ODE. For medical image segmentation, LSS-SKAN-UNet enhances F1 scores by 2.00\% while utilizing fewer parameters. In the course of this analysis, our findings also yield the first strong empirical support for the theoretical proposition that basis function smoothness is essential for stable KAN training \cite{samadi2024smooth}. Prior to this, article \cite{samadi2024smooth} only gave a theoretical analysis.
\end{enumerate}

These contributions provide a cohesive theoretical framework, a guiding design philosophy, and a practical, high-performance model family, collectively charting a new direction for efficient neural network design.

\section{Related Work}

\textbf{Theoretical Foundation.}
The Kolmogorov-Arnold representation theorem (KART) is the theoretical cornerstone of KAN architectures \cite{somvanshi2024surveykolmogorovarnoldnetwork,suh2024survey}.
This theorem posits that continuous multivariate functions can be expressed as compositions of univariate functions, with Arnold simplifying Kolmogorov's original proof \cite{kolmogorov1956representation}.
This mathematical principle underpins KAN's design for representing high-dimensional functions.
The analogies between neural networks and KART have fueled theoretical research \cite{suh2024survey,poggio1989theory,girosi1989representation,schmidt2021kolmogorov}, with recent studies further establishing an equivalence between smooth KANs and MLPs for specific function classes \cite{samadi2024smooth}.

\textbf{Development of KAN Variants.}
The original KAN architecture \cite{liu_kan_2024} utilized B-splines as learnable activation functions.
Subsequently, numerous KAN variants have been developed, including Wav-KAN \cite{bozorgasl_wav-kan_2024} employs wavelet functions, Fast-KAN \cite{li_kolmogorov-arnold_2024} uses Gaussian radial basis functions for faster forward propagation, and FourierKAN \cite{FourierKAN} incorporates Fourier series.
Other variants include fKAN~\cite{aghaei_fkan_2024} and rKAN \cite{aghaei_rkan_2024} with adaptive fractional orthogonal Jacobi functions and Padé approximations with rational Jacobi functions, respectively. These diverse basis function designs are primary benchmarks for SKANs.
In addition, KANs have demonstrated applicability across various domains, including composite networks \cite{cheon_kolmogorov-arnold_2024}, graph-based tasks \cite{xu_fourierkan-gcf_2024,kiamari_gkan_2024,de_carlo_kolmogorov-arnold_2024,bresson_kagnns_2024}, time series analysis \cite{genet_tkan_2024,vaca-rubio_kolmogorov-arnold_2024}, and diverse scientific computing problems \cite{liu_kan_2024-1,howard_finite_2024,wang_kolmogorov_2024,abueidda_deepokan_2024,kashefi2024kolmogorov}.
In this work, we specifically validates SKANs in differential equation solving and medical image segmentation.

\textbf{KAN in Differential Equation Solving.}
The Neural ODE framework \cite{chen2018neural} significantly advanced the use of neural networks for solving ODEs.
This was extended by works incorporating physical constraints \cite{raissi2019physics}, invariants \cite{cranmer2020lagrangian}, and symplectic properties \cite{jin2020sympnets}.
KAN-ODEs \cite{Koenig_2024} were later introduced, integrating KANs for higher precision than MLP-based Neural ODEs. 

\textbf{KAN in Medical Image Segmentation.}
U-Net \cite{ronneberger2015u} is a foundational deep learning architecture for medical image segmentation.
Many enhancements like U-Net++ \cite{zhou2018unet++}, ResU-Net++ \cite{jha2019resunetadvancedarchitecturemedical}, Attention U-Net \cite{oktay2018attention}, and nnU-Net \cite{isensee2021nnu} primarily use MLPs, which can restrict modeling capacity.
U-KAN \cite{Li_Liu_Li_Wang_Liu_Liu_Chen_Yuan_2025} integrated Spl-KAN into U-Net and  exhibited improved performance than MLP-based models. 

\section{Rethinking KAN Fundamentals: The Case for Architectural Scaling}
\label{sec:rethinking_kan_fundamentals_final_pure}

\vspace{2mm}
\textbf{Limitations of an Overemphasis on Basis Function Complexity in KANs.}
The vanilla KAN architecture, specifically Spl-KAN, introduced learnable B-spline basis functions, a significant departure from the fixed activations of traditional MLPs~\cite{liu_kan_2024}.
The parameterization of these B-splines, particularly the grid size ($g$), directly determines their complexity and the number of learnable parameters per basis function:

\vspace{-2mm}
\begin{equation}
N_{p}^{Spl} = \sum\limits_{l=0}^{L-1} \underbrace{n_{in}^{(l)} \cdot n_{out}^{(l)}}_{\substack{\text{layer connectivity}}} \cdot \underbrace{(g + s_{o} + 1)}_{\substack{\text{basis complexity}}}
\end{equation}
where $L$ denotes network depth, $n_{in/out}^{(l)}$ specify input/output dimensions at layer $l$, $g$ controls spline grid resolution, and $s_{o}$ is the polynomial spline order.
Subsequent research, while diversifying basis function types (e.g., wavelets~\cite{bozorgasl_wav-kan_2024}, radial basis functions~\cite{li_kolmogorov-arnold_2024}, Fourier series~\cite{FourierKAN}), has largely perpetuated a focus on the intrinsic sophistication of these univariate functions~\cite{aghaei_fkan_2024, aghaei_rkan_2024}.

However, this pursuit of highly expressive individual basis functions, while intuitive, presents several theoretical and practical drawbacks. Firstly, it can lead to \textbf{suboptimal parameter allocation}.
Increasing parameters per basis function (e.g., larger $g$) inherently limits network scale (depth/width) under a fixed total parameter budget.
This trade-off is critical because the ability of a network to model complex, high-dimensional functions often relies on the interactions learned across a broad and deep architecture than on the isolated power of its individual components. The KART itself guarantees representation with univariate functions, but it does not prescribe their complexity for optimal \textit{learning} or \textit{efficiency} in a finite-data, finite-computation regime.

Secondly, complex basis functions can exacerbate \textbf{optimization difficulties}.
A higher number of parameters per basis can lead to more convoluted local loss landscapes, potentially increasing the likelihood of converging to less optimal solutions or requiring more intricate optimization strategies. Simpler functions, conversely, may offer smoother optimization pathways.

Thirdly, the \textbf{interpretability of KAN can be challenged by basis function complexity}. While KANs, including Spl-KANs, offer a transparent network structure where each activation pathway is, in principle, inspectable, a distinct advantage over the opaque internal workings of MLPs, this benefit hinges on the comprehensibility of the individual basis functions. The ideal scenario for Spl-KAN interpretability assumes its splines can accurately capture and represent true underlying univariate functions in an analyzable form. However, this is not guaranteed; the learned splines might not always correspond to simple, symbolic expressions. More critically, if each basis function itself becomes an intricate, multi-parameter entity (e.g., a B-spline with many knots), it can devolve into a miniature black box. Understanding the behavior of such a complex learned spline can become as challenging as interpreting a small sub-network, thereby diluting the intended clarity. Effective interpretability in KANs is thus best served when the individual functional components remain transparent and relatively simple to analyze. 
These considerations suggest a need to re-evaluate the balance between basis function complexity and overall network architecture.

\vspace{2mm}
\textbf{The Efficient KAN Expansion (EKE) Hypothesis.}
Based on the limitations discussed, we propose a shift in the design philosophy for KANs, which we formulate as the Efficient KAN Expansion (EKE) hypothesis. This hypothesis posits that the expressive power and learning efficiency of a KAN, under a constrained parameter budget, are better served by increasing the network's architectural scale using simpler, less parameterized basis functions.

\begin{mdframed}
	\textit{Hypothesis 1 (Efficient KAN Expansion - EKE):}
	Let the total parameters of a KAN be $N_{p}=N_c \cdot N_b$, where
	$N_c$ quantifies network topology complexity and 
	$N_b$ captures basis function complexity. For example, $N_c = \textstyle\sum_l n_{in}^{(l)}n_{out}^{(l)}$ and $N_b = g + s_{o} + 1$ in Spl-KAN. Given a fixed $N_{p}$, we hypothesize that architectures achieving higher $N_c$ through deeper topology or wider layers while maintaining minimal viable $N_b$, will exhibit superior model performance. Formally, 
	\begin{equation}
		\mathcal{M}\left(N_c, N_b\right) \geq \mathcal{M}\left(N_c^{\prime}, N_b^{\prime}\right)    
	\end{equation}
	when $N_b \leq N_b^{\prime}$ and the saved parameters enable  $N_c \geq N_c^{\prime}$ while maintaining $N_c \cdot N_b \approx N_c^{\prime} \cdot N_b^{\prime}$.
\end{mdframed}

Several theoretical considerations underpin the EKE hypothesis. Primarily, prioritizing architectural scale (depth and width) enhances the network's expressive power through multiple synergistic effects. A larger network allows for \textit{richer distributed representations}, where numerous simpler basis functions collectively model complex relationships more robustly than a few intricate ones. This scale also facilitates deeper \textit{hierarchical feature abstraction}, a key benefit often limited when parameters are heavily invested in basis function complexity rather than network layers. Furthermore, the \textit{combinatorial expressiveness} grows significantly with scale, as a larger number of simple transformations can be combined in vastly more ways than fewer complex ones. From a learning perspective, simpler basis functions generally lead to a more \textit{manageable optimization landscape} with potentially smoother gradients and improved training stability, enhancing overall learnability. This philosophy aligns with successful paradigms in other domains, like CNNs achieving power through compositions of simple filters, suggesting that complexity arising from the \textit{scaled composition of simple elements} is a potent and efficient strategy \cite{fukushima_neocognitron_1980}.


\vspace{2mm}
\textbf{Empirical Validation of the EKE Hypothesis.}
\begin{figure*}[!h]
	\centering
	\includegraphics[width=0.9\textwidth]{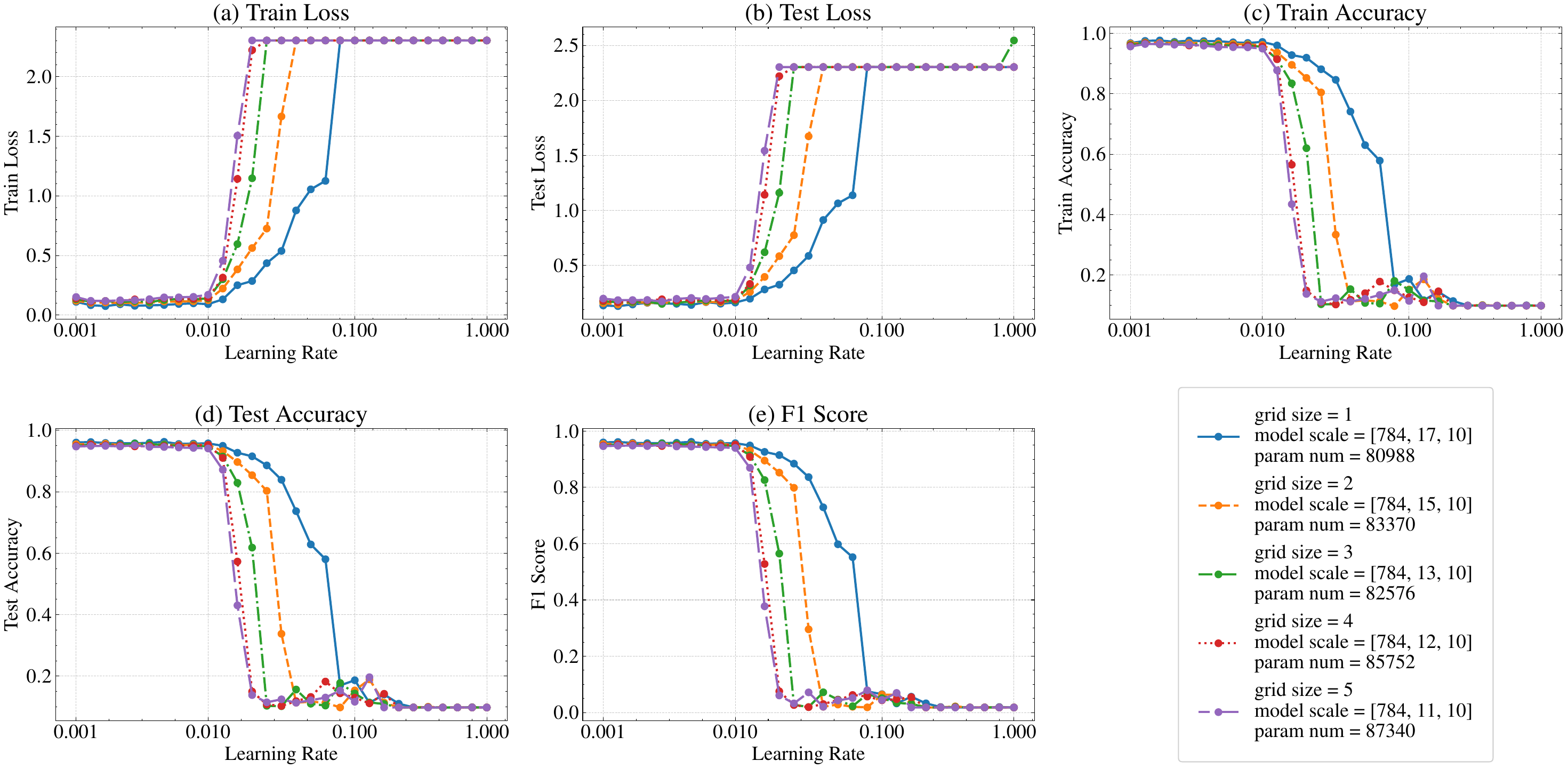}
	\caption{Comprehensive performance evaluation of Spl-KAN with varying grid sizes across a wide spectrum of learning rates. The analysis encompasses: (a) training loss, (b) test loss, (c) training accuracy, (d) test accuracy, and (e) F1 score. The results consistently identify an optimal performance window within the [0.001, 0.01] learning rate range, guiding subsequent focused analysis.} 
	\label{fig:spl_kan_full_lr}
\end{figure*}
\begin{figure*}[!h]
	\centering
	\includegraphics[width=0.9\textwidth]{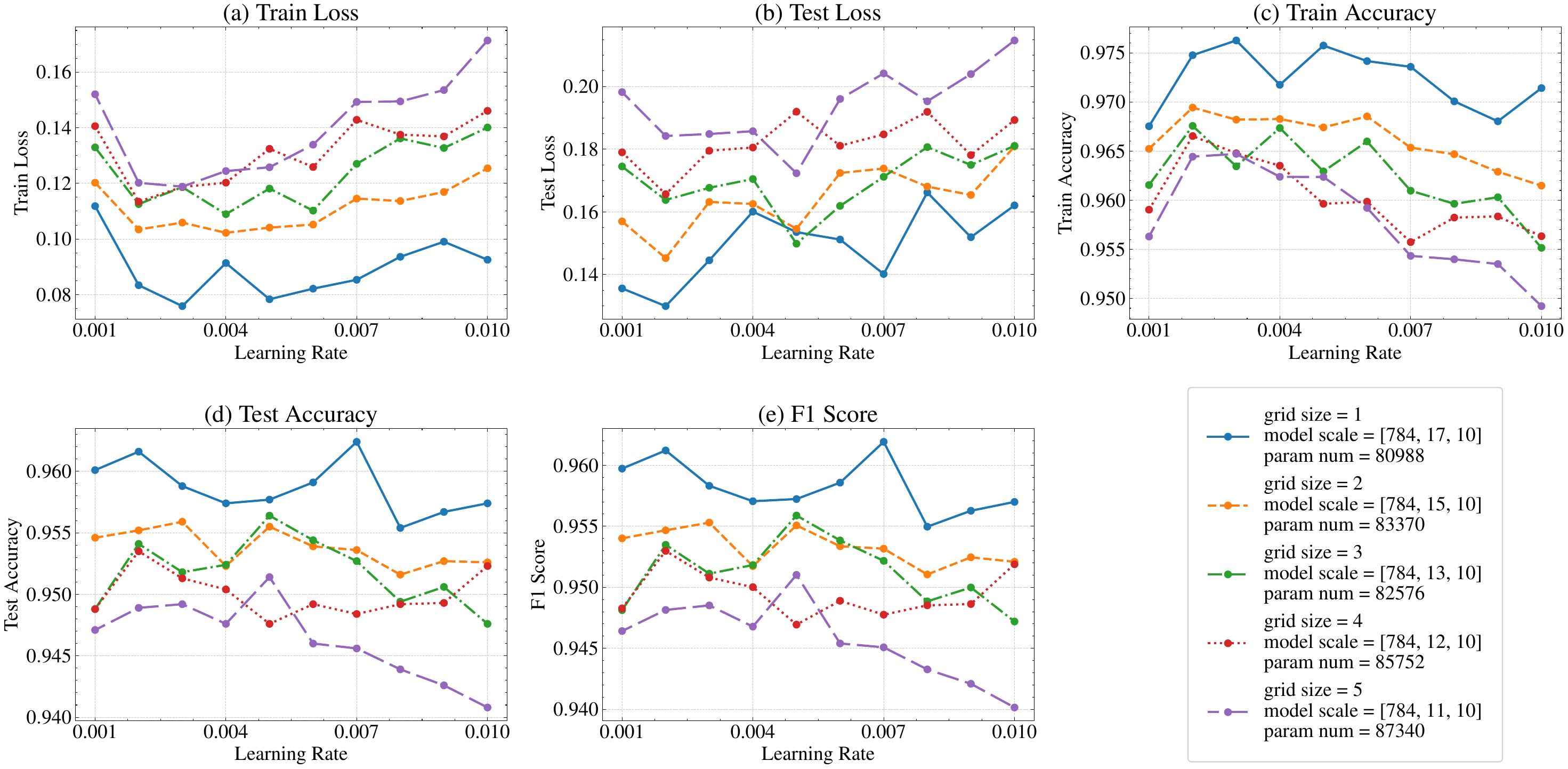}
	\caption{Detailed performance analysis within the optimal learning rate range [0.001, 0.01]. For visual clarity, only grid sizes 1 through 5 are shown. These plots reveal a systematic performance advantage for smaller grid sizes (e.g., g=1, g=2) across all five key metrics, providing strong evidence for the EKE hypothesis.} 
	\label{fig:spl_kan_focused_lr}
\end{figure*}
\begin{figure}[!h]
	\centering
	\includegraphics[width=0.45\textwidth]{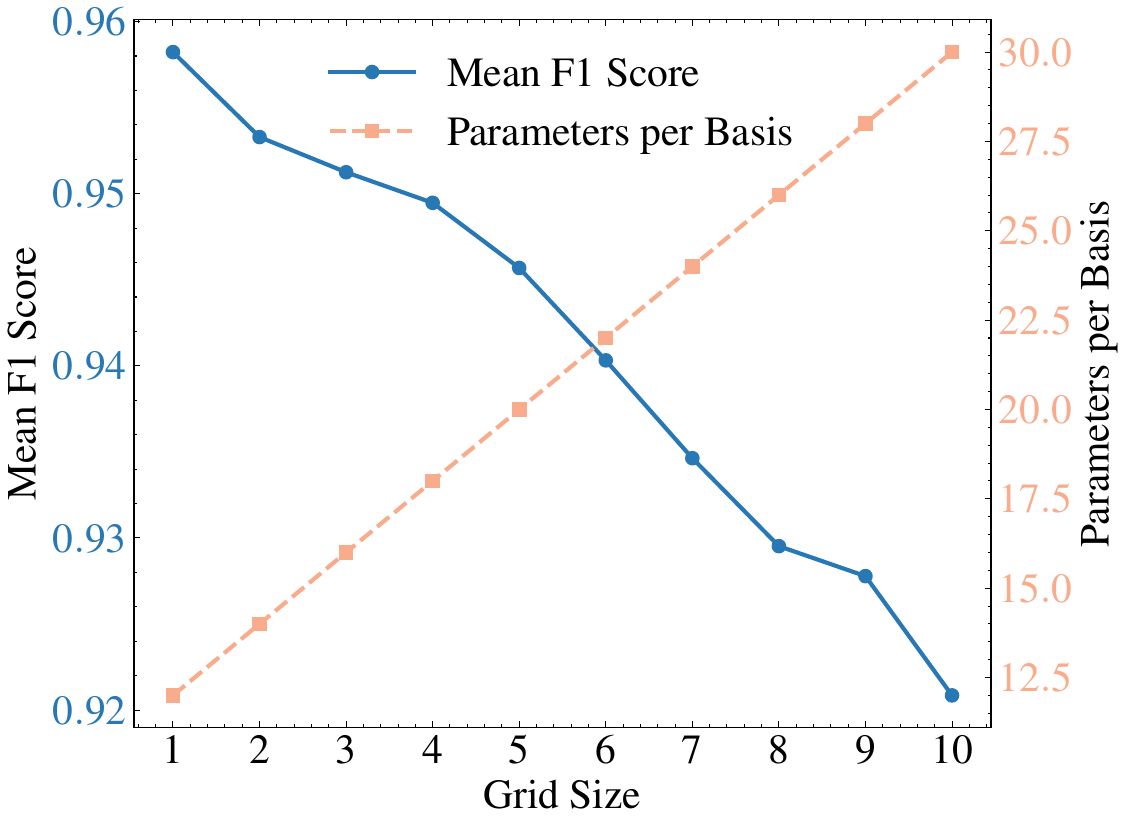}
	\caption{The trade-off between basis function complexity and model performance in Spl-KAN, evaluated for grid sizes 1 through 10. The inverse correlation between mean F1 score and grid size, juxtaposed with the direct correlation between basis parameters and grid size, offers a compelling empirical validation of the EKE hypothesis.} 
	\label{fig:spl_kan_grid_analysis}
\end{figure}
To establish a rigorous empirical foundation for the EKE hypothesis, we conducted a controlled experiment designed to meticulously isolate the trade-off between basis function complexity and architectural scale. Using the original Spline-KAN (Spl-KAN) on the MNIST dataset, we enforced a parameter budget of approximately 80,000 across all configurations. The pivotal variable was the B-spline grid size ($g$), which directly governs the complexity of each basis function. We systematically evaluated configurations with grid sizes ranging from $g=1$ (minimal complexity) to $g=10$. To ensure robustness, each configuration was trained over 10 independent runs, with an extensive linear search for the optimal learning rate across key intervals. The results presented are the best-performing outcomes for each model configuration.

Our initial analysis spanned a broad spectrum of learning rates to map the optimization landscape. For the sake of drawing beauty, we only draw 5 curves with grid size from 1 to 5 in \figurename~\ref{fig:spl_kan_full_lr} and \figurename~\ref{fig:spl_kan_focused_lr}. As illustrated in \figurename~\ref{fig:spl_kan_full_lr}, this comprehensive sweep reveals a critical insight: all model configurations, irrespective of grid size, exhibit a stable and effective performance profile predominantly within the [0.001, 0.01] learning rate interval. Beyond this window, particularly for learning rates exceeding 0.01, performance degrades precipitously, indicating training instability. This observation not only identifies the optimal operational range for these architectures but also justifies a more granular analysis within this specific interval.

Within this optimal range, an in-depth analysis exposes a striking and unequivocal trend, as detailed in \figurename~\ref{fig:spl_kan_focused_lr}. A clear and systematic performance hierarchy emerges, where architectures employing simpler basis functions (e.g., $g=1, g=2$) consistently and substantially outperform those with more complex bases ($g=4, g=5$). This dominance is not confined to a single metric but is evident across the entire suite of evaluation criteria—training/test loss, training/test accuracy, and F1 score. Given the fixed parameter budget, utilizing simpler bases liberates resources that are reallocated to expand the network's width and depth (as detailed in the figure legends). This outcome strongly validates the central tenet of the EKE hypothesis: the performance gains from architectural scaling far outweigh any marginal benefits from increasing the expressive power of individual basis functions.

To crystallize this trade-off and demonstrate its persistence, \figurename~\ref{fig:spl_kan_grid_analysis} presents a complete visualization of the relationship between basis function complexity and model performance, extending the analysis to a grid size of 10. The plot juxtaposes the linear increase in parameters per basis function against the resulting mean F1 score within the optimal learning rate window. The two curves delineate a stark inverse correlation: as basis complexity methodically increases, performance systematically degrades. This is not an ambiguous or noisy relationship but a clear, monotonic decline. This visualization provides decisive empirical evidence for the EKE hypothesis, demonstrating that under parameter constraints, sacrificing network scale for basis function complexity is a demonstrably suboptimal design choice. These findings serve as the foundational motivation for the SKAN architecture.


\section{The Uni-KAN Framework: A Unified Abstraction for KANs}

The rapid proliferation of KAN variants necessitates a unified architectural abstraction. To this end, we introduce the \textbf{Universal KAN (Uni-KAN)} framework, a generalized structure capable of representing most KAN-style networks. This framework not only provides a common descriptive language, but is also designed with computational efficiency and hardware acceleration in mind.

\begin{figure}[t]
	\centering
	\includegraphics[width=0.9\columnwidth]{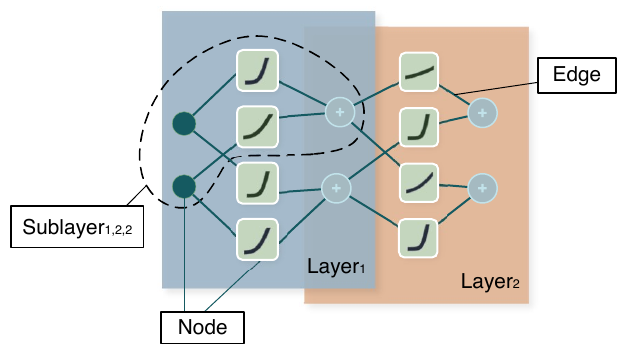}
	\caption{Architectural primitives of the Uni-KAN framework. The network consists of layers formed by nodes and edges. We define the \textbf{sublayer} (e.g., $S_{1,2:2}$, circled) as the minimal computational unit, fundamental for parallelized computation.}
	\label{fig:uni-kan-primitives}
\end{figure}

\subsection{Formal Definition of a Generalized KAN}

We first formalize the structure of a KAN-style network.

\begin{definition}[Generalized KAN]
	A Generalized KAN is a directed acyclic graph organized into layers $l \in \{0, \dots, L-1\}$. A layer $l$ performs a transformation $\varPhi^{(l)}: \mathbb{R}^{n_{in}^{(l)}} \to \mathbb{R}^{n_{out}^{(l)}}$. The output $y_j^{(l)}$ of the $j$-th node in layer $l$ is computed by an aggregation operation $\bigoplus$ over transformations applied to inputs from the previous layer:
	\begin{equation}
		y_j^{(l)} = \bigoplus_{i=1}^{n_{in}^{(l)}} \phi_{j,i}^{(l)}(x_i^{(l-1)})
	\end{equation}
	where $\phi_{j,i}^{(l)}$ is a learnable univariate basis function on the edge connecting input node $i$ to output node $j$. In this work, we focus on summation nodes, where $\bigoplus = \sum$.
\end{definition}

Within this structure, we distinguish between two representational schemes based on their connectivity and basis function constraints.

\begin{definition}[Sparse and Dense KANs]
	A KAN is defined as \textbf{Sparse} if the basis functions $\phi_{j,i}^{(l)}$ can be chosen arbitrarily and independently for each edge. A KAN is defined as \textbf{Dense} if for any given output node $j$, all its incoming basis functions $\{\phi_{j,i}^{(l)}\}_{i=1}^{n_{in}^{(l)}}$ are of the same functional form, differing only in their learnable parameters.
\end{definition}

The Sparse representation offers maximum flexibility, while the Dense representation is structured for efficient, vectorized computation. One of the key insights of our framework is that these two are formally equivalent.

\subsection{Representation Equivalence}

\begin{figure}[t]
	\centering
	\includegraphics[width=\columnwidth]{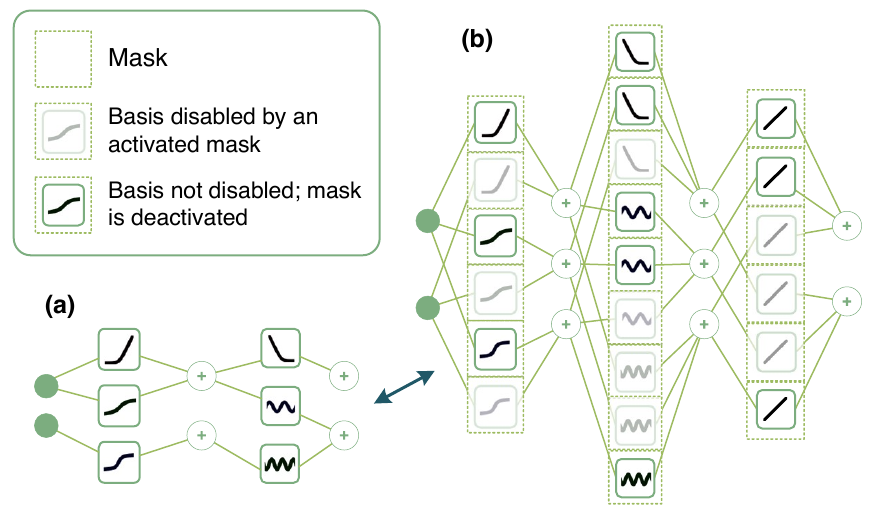}
	\caption{Interchangeability of Sparse and Dense Uni-KAN. (a) A Sparse KAN. (b) Its equivalent Dense KAN representation, using a binary \textbf{mask} to deactivate edges (faded), replicating the sparse topology within a regular format.}
	\label{fig:uni-kan-sparse-dense}
\end{figure}

The core strength of the Uni-KAN framework is its generality, providing a single, coherent structure to define most KAN variant. A powerful consequence of this unified design is the ability to losslessly represent a flexible Sparse KAN in a Dense format suitable for hardware acceleration. We formalize this interchangeability as a proposition.

\begin{proposition}[Uni-KAN Representation Equivalence]
	Any Sparse KAN, $\mathcal{K}_{sparse}$, with an arbitrary topology can be exactly represented by a Dense KAN, $\mathcal{K}_{dense}$, paired with a static binary mask, $M$.
\end{proposition}
\begin{proof}[Proof Sketch]
	Let $\mathcal{K}_{sparse}$ have an arbitrary connection topology. We can construct a corresponding $\mathcal{K}_{dense}$ with the same layer dimensions, where each output node $j$ in layer $l$ is fully connected to all input nodes $i$ via a default basis function $\phi_{j,*}^{(l)}$. We then introduce a binary mask tensor $M^{(l)}$ of shape $(n_{out}^{(l)}, n_{in}^{(l)})$ where $M_{j,i}^{(l)}=1$ if an edge exists from $i$ to $j$ in $\mathcal{K}_{sparse}$, and $M_{j,i}^{(l)}=0$ otherwise. The output of the equivalent layer is then:
	\begin{equation}
		y_j^{(l)} = \sum_{i=1}^{n_{in}^{(l)}} M_{j,i}^{(l)} \cdot \phi_{j,i}^{(l)}(x_i^{(l-1)})
	\end{equation}
	This operation, typically an element-wise product, effectively deactivates the non-existent edges, as shown in \figurename~\ref{fig:uni-kan-sparse-dense}. This construction demonstrates that any sparse topology can be embedded within a dense computational graph, unifying architectural flexibility with computational efficiency. For cases where $\mathcal{K}_{sparse}$ uses heterogeneous basis functions, a composite $\mathcal{K}_{dense}$ can be formed by layering multiple dense sub-layers, each with its own mask.
\end{proof}

This Uni-KAN framework provides the theoretical and practical foundation for our work. SKANs, which we introduce next, can be understood as a highly effective realization of the Dense Uni-KAN paradigm, designed in accordance with the EKE hypothesis to maximize both architectural and computational efficiency.


\section{SKANs: Single-Parameter KANs}
The Uni-KAN framework provides the blueprint for constructing arbitrary KANs. We now introduce SKANs, a specific family of KANs that optimally embodies the EKE hypothesis.

\textbf{Definition of SKANs.}
Based on the EKE hypothesis, which emphasizes scaling network size over increasing basis function complexity for improved performance, we propose a streamlined architecture: Single-Parameter KANs (SKANs). In SKANs, each basis function is designed with only a single learnable parameter, facilitating more efficient network construction. 
The visualization in \figurename~\ref{fig:overview}(b) presents a [2,2,2] SKAN architecture, which 
follows the general KAN framework, where each edge represents a parameterized function, while the intermediate and final nodes perform addition operations exclusively. The distinguishing feature of this design lies in the edge functions, which contain only one learnable parameter.

Formally, an SKAN is defined as:
\begin{equation}
S K A N(x)=\left(\varPhi^{(L-1)} \circ \varPhi^{(L-2)} \circ \cdots \circ \varPhi^{(0)}\right)(x)
\end{equation}
where each layer $\varPhi^{(l)}$ is a matrix of functions:
%
\begin{equation}
\varPhi^{(l)}=\left[\begin{array}{cccc}
\phi_{1,1}^{(l)} & \phi_{1,2}^{(l)} & \cdots & \phi_{1, n_{i n}^{(l)}}^{(l)} \\
\phi_{2,1}^{(l)} & \phi_{2,2}^{(l)} & \cdots & \phi_{2, n_{i n}^{(l)}}^{(l)} \\
\vdots & \vdots & \ddots & \vdots \\
\phi_{n_{o u t}^{(l)}, 1}^{(l)} & \phi_{n_{o u t}^{(l)}, 2}^{(l)} & \cdots & \phi_{n_{o u t}^{(l)}, n_{i n}^{(l)}}^{(l)}
\end{array}\right]
\end{equation}
Here, $\phi_{i, j}^{(l)}(k_{i, j}^{(l)}, x_j)$ is a learnable basis function connecting the $j$-th neuron of layer $l-1$ to the $i$-th neuron of layer $l$, containing only a single learnable parameter $k_{i, j}^{(l)}$. The outputs of these functions are summed at each node in the subsequent layer.

\textbf{Systematic Design of Single-Parameter Basis Functions.}
The design of effective single-parameter basis functions is crucial for SKANs. Our approach involved a systematic review and adaptation of 1) common activation functions from deep learning and 2) periodic functions inspired by the Fourier transform theory. 

\textit{1) Design Based on Classical Activation Functions.} We transformed well-established activation functions into learnable single-parameter basis functions by introducing a parameter $k$ that modulates their behavior while preserving their core characteristics. The naming convention adopts a prefix ``L'' to signify that these functions are learnable variants, followed by the name of their prototype counterpart functions. The designed functions include:

a) ReLU family
\begin{align}
	\text{LReLU: } &f(k,x)= \max(kx,0) \\
	\text{LLeakyReLU: } &f(k,x)= \max(kx,x) \\
	\text{LHardSigmoid: } &f(k,x) = \max(0, \min(1, kx + 0.5))
\end{align}

b) Sigmoidal variants
\begin{align}
	\text{LSwish: } &f(k,x)= \frac{x}{1 + e^{-kx}} \\
	\text{LMish: } &f(k,x) = x \cdot \tanh(\ln(1 + e^{kx})) \\
	\text{LGELU: } &f(k,x) = 0.5x(1 + \tanh(\sqrt{\frac{2}{\pi}}(kx \nonumber \\ 
	& \qquad \qquad + 0.044715k^3x^3)))
\end{align}

c) Smooth saturations
\begin{align}
	\text{LELU: } &f(k,x) = \begin{cases} k \cdot (e^{x/k} - 1) & \text{if } x < 0 \\ x & \text{otherwise} \end{cases} \\
	\text{LSoftplus: } &f(k,x)= \ln(1+e^{kx}) \\
	\text{LSS: } &f(k,x) = \ln(1 + e^{kx}) - \ln(2)
\end{align}












\begin{figure*}[!h]
	\centering
	\includegraphics[width=\textwidth]{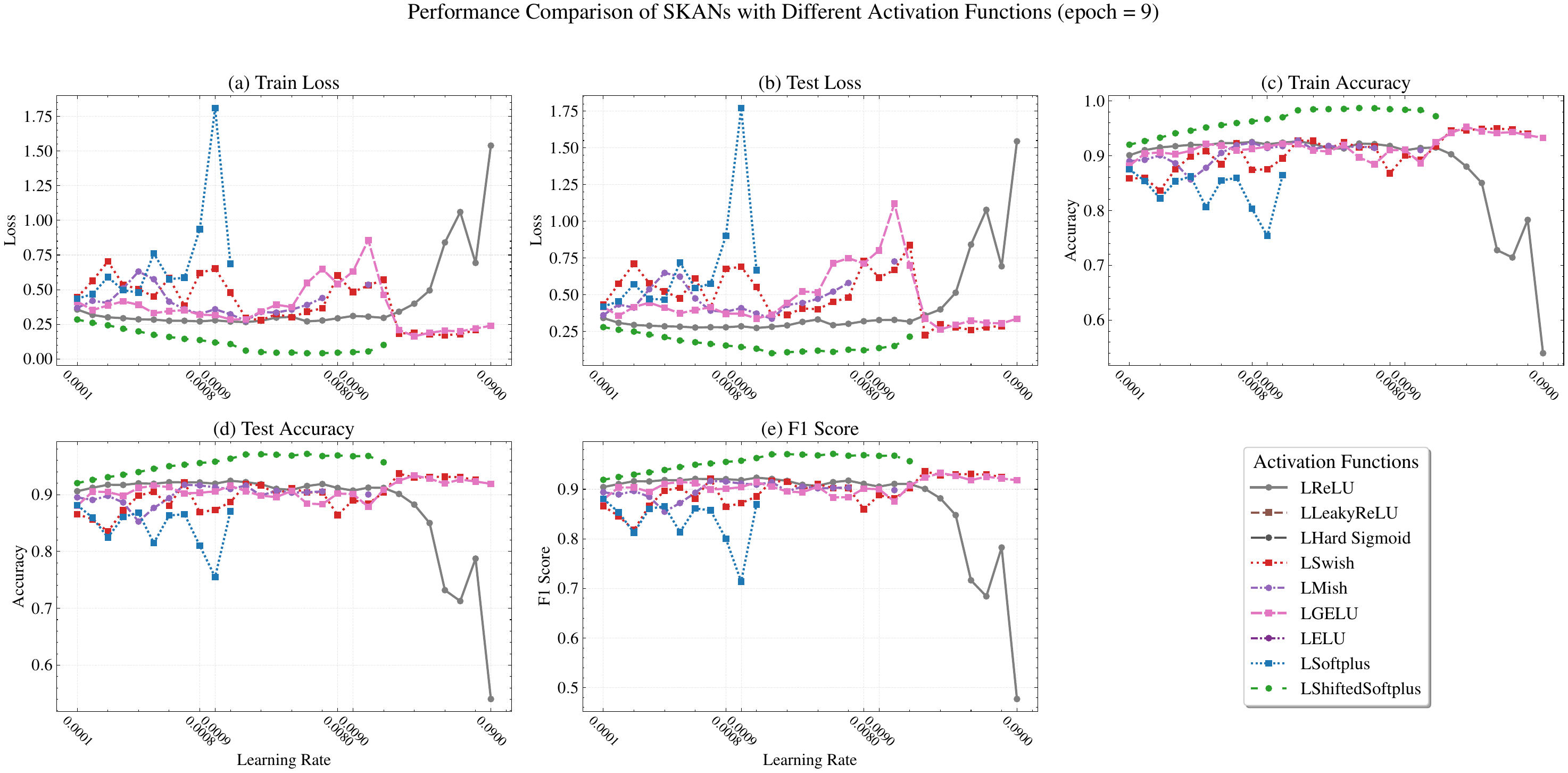} 
	\caption{Performance comparison of SKAN variants across a wide learning rate spectrum on the MNIST dataset (epoch=9). Subplots show (a) Train Loss, (b) Test Loss, (c) Train Accuracy, (d) Test Accuracy, and (e) F1 Score. The results highlight a general trend of instability at higher learning rates and identify an optimal performance window between 0.001 and 0.01.}
	\label{fig:skan_comparison_full}
\end{figure*}

\begin{figure*}[!h]
	\centering
	\includegraphics[width=\textwidth]{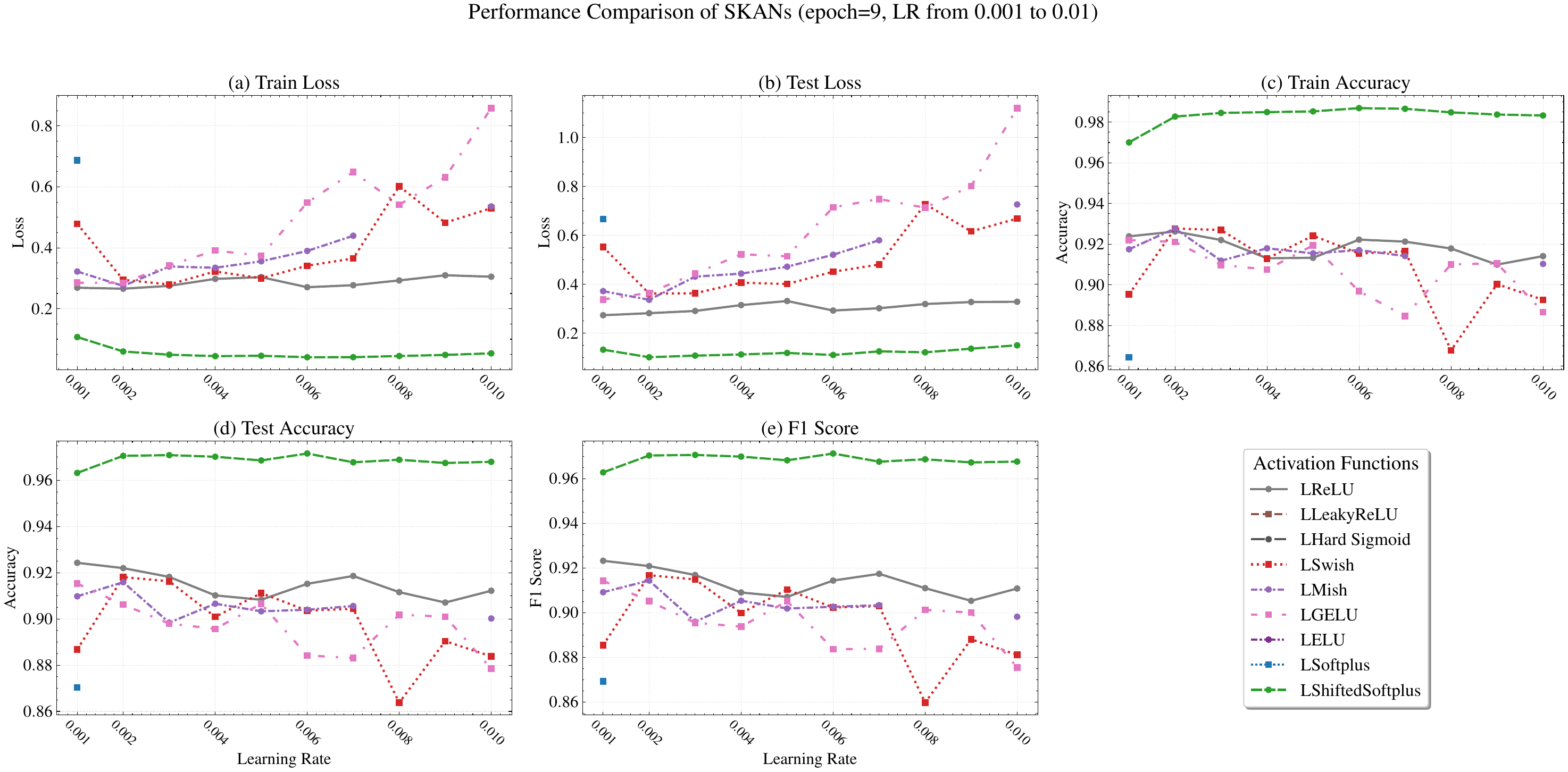} 
	\caption{A focused view of SKAN performance within the optimal learning rate interval [0.001, 0.01]. In this range, the Learnable Shifted Softplus (LSS) variant (green, dash-dotted line) demonstrates clear superiority across all key metrics, achieving the lowest loss and highest accuracy/F1 score, while other functions exhibit greater variability.}
	\label{fig:skan_comparison_focused}
\end{figure*}

\begin{table*}[!h]
	\centering
	\caption{Performance Summary of SKAN Variants with Classical Activation-Based Functions on MNIST (10 Epochs)}
	\label{tab:activation_performance_summary}
	\begin{adjustbox}{max width=\textwidth}
		\small
		\setlength{\tabcolsep}{3pt}
		\begin{tabular}{l|cccccccc|c}
			\toprule
			Metric & LReLU & LLeakyReLU & LHardSigmoid & LSwish & LMish & LGELU & LELU & LSoftplus & LSS \\
			\midrule
			Best F1 Score & 0.9232 & \textit{Unstable} & \textit{Unstable} & 0.9358 & 0.9166 & 0.9328 & \textit{Unstable} & 0.8808 & \textbf{0.9713} \\
			Avg Runtime (s) & \textbf{5.0570} & \textit{N/A} & \textit{N/A} & 5.7955 & 5.7301 & 6.4208 & \textit{N/A} & 5.3071 & 6.0093 \\
			\bottomrule
		\end{tabular}
	\end{adjustbox}
\end{table*}

To adjudicate between these candidate functions, we conducted a rigorous empirical evaluation on the MNIST dataset. All SKAN variants were instantiated with a consistent architecture and trained for 10 epochs. A comprehensive linear search across a wide range of learning rates was performed to map the performance landscape for each function.

The results, presented in \figurename~\ref{fig:skan_comparison_full}, reveal critical insights into the performance and stability of different basis functions. The panoramic view across all learning rates shows that while most functions perform reasonably at lower rates, they are highly susceptible to training instability as the learning rate increases, often leading to performance collapse. This initial sweep allowed us to identify an optimal and stable learning rate window, specifically the $[0.001, 0.01]$ interval.

A focused analysis within this optimal window, detailed in \figurename~\ref{fig:skan_comparison_focused}, exposes a clear performance hierarchy.
\begin{enumerate}[leftmargin=*]
	\item \textbf{Superior Performance of LSS:} The \textbf{Learnable Shifted Softplus (LSS)} variant emerges as the unequivocal top performer. It consistently achieves the lowest test loss and the highest F1 score (approaching 0.97), significantly outperforming all other candidates. Its stable and dominant performance across the optimal learning rate range highlights its robustness and effectiveness.
	
	\item \textbf{Instability of Non-Smooth Functions:} A crucial observation is the pronounced training instability of several functions, particularly LLeakyReLU, LHardSigmoid, and LELU, as corroborated in Table~\ref{tab:activation_performance_summary}. These functions are characterized by piecewise linear structures and points of non-differentiability (i.e., they are not smooth). Their volatile performance, marked by sharp fluctuations in loss and accuracy, provides strong empirical evidence for the hypothesis that \textbf{basis function smoothness is paramount for stable KAN training}. This finding aligns perfectly with recent theoretical work~\cite{samadi2024smooth}, which suggests that smooth activation pathways are essential for ensuring stable gradient flow and reliable convergence in KAN-style networks. Beyond demonstrating the superiority of LSS, this analysis also serves as a quantitative validation of the notion that smoothness is a key prerequisite for stable KAN training, as posited in theory by \cite{samadi2024smooth}. The continuous differentiability of LSS is a likely contributor to its superior stability.
\end{enumerate}

Table~\ref{tab:activation_performance_summary} gives the best F1 score and average running time of each network at all learning rates. The numerical pattern of the best F1 score is consistent with the trend in \figurename~\ref{fig:skan_comparison_focused}, while LRELU has the fastest running time, but the training speed of other stable trainable networks (including LSS) is also similar. Based on its exceptional synthesis of accuracy, stability, and computational efficiency (Table~\ref{tab:activation_performance_summary}), \textbf{LSS was selected as the primary basis function for general-purpose SKAN applications}.

\vspace{1mm} 
\textit{2) Design Based on Periodic Functions.} Inspired by Fourier theory, we also developed trigonometric-based periodic functions by introducing a learnable parameter $k$ to scale the input $x$:
\begin{align}
\text{LSin: } f(k,x) &= \sin(kx) \\ 
\text{LCos: } f(k,x) &= \cos(kx) \\
\text{LArctan: } f(k,x) &= \arctan(kx) 
\end{align}


These functions, particularly LArctan and LSin, also demonstrate strong performance in various tasks, as detailed in Section~\ref{sec:experiments}.

\textbf{Analysis of Gradient Stability.}
The choice of basis function profoundly impacts the stability of gradients during training. Complex basis functions, especially those with many parameters or highly non-linear regions, can be prone to issues like vanishing or exploding gradients. Single-parameter functions, by their nature, tend to have simpler derivative forms.
For instance, functions like LReLU and LLeakyReLU, while simple, can sometimes suffer from issues related to zero gradients (in the case of LReLU for negative inputs). Our empirical findings showed that LELU, LLeakyReLU, and LHardSigmoid exhibited training instability, potentially due to challenges in maintaining stable gradient flow across layers and varying inputs.
In contrast, functions like LSS, which is a smoothed, shifted version of ReLU, demonstrated superior stability and accuracy. The smoothness of LSS is crucial for KAN design \cite{samadi2024smooth}, and the property that it passes through the origin can enable better gradient flow during training. Similarly, trigonometric functions like LSin, LCos, and LArctan have bounded derivatives (for LSin and LCos, the derivatives are also trigonometric functions, and for LArctan, the derivative $\frac{k}{1+(kx)^2}$ is bounded and diminishes for large $|kx|$), which can contribute to more controlled gradient propagation. The single learnable parameter $k$ primarily scales the input, leading to predictable changes in the function's behavior and its derivative, thus simplifying the optimization landscape compared to multi-parameter basis functions.

\textbf{Integration with Neural Network Architectures.}
With the SKAN architecture and its basis functions defined, integrating these components into practical neural network applications becomes essential to validate the effectiveness of SKAN design in specific scenarios such as differential equation solving and medical image segmentation. The integration process needs to preserve the efficiency of single-parameter design.

The architectural integration of KAN variants into existing neural networks follows a consistent pattern: architectures in \cite{Koenig_2024} and \cite{Li_Liu_Li_Wang_Liu_Liu_Chen_Yuan_2025} are derived by replacing the MLP components in Neural ODE \cite{chen2018neural} and U-Net \cite{ronneberger2015u} with Spl-KAN, respectively. Following this established approach, we extend our SKANs to SKAN-ODE and SKANs-UNet architectures, which are constructed by substituting the Spl-KAN components in \cite{Koenig_2024} and \cite{Li_Liu_Li_Wang_Liu_Liu_Chen_Yuan_2025} with the proposed SKAN architecture. The nomenclature for specific implementations follows a systematic pattern: when a particular basis function is employed, the corresponding models are denoted as x-SKAN-ODE and x-SKAN-UNet (e.g., LSS-SKAN-ODE and LSS-SKAN-UNet for implementations using the LSS basis function).

\section{Experiments}
\label{sec:experiments}
This study systematically addresses three research questions:
\textbf{RQ1}: What is the impact of single-parameter basis functions (LSS, LArctan, LSin) in SKANs on MNIST classification accuracy and computational efficiency, and how do SKANs benchmark against existing KAN variants and conventional baselines? 
\textbf{RQ2}: Can SKANs achieve competitive performance in specialized tasks? We validate SKAN-ODE on differential equation solving and SKANs-UNet on medical image segmentation against domain-specific baselines and KAN derivatives, emphasizing parameter efficiency.
\textbf{RQ3}: Does cross-domain evidence from the above experiments (MNIST classification, ODE solving, segmentation) substantiate the EKE hypothesis proposed in Section~\ref{sec:rethinking_kan_fundamentals_final_pure}?
The section first outlines experimental configurations, then presents structured results corresponding to the three questions.

\begin{table*}[]
\centering
\caption{Performance Comparison of Different Methods on MNIST}
\label{tab:performance_comparison_single_row_adjusted}
\begin{adjustbox}{max width=\textwidth} 
\small 
\setlength{\tabcolsep}{3pt} 
\begin{tabular}{l|cccccc|ccc}
\toprule
Metric & Spl-KAN & WavKAN & FastKAN & FourierKAN & MLP+fKAN & MLP+rKAN & LSS-SKAN & LArctan-SKAN & LSin-SKAN \\
\midrule
Best F1 Score   & 0.9615  & 0.9754 & 0.9515  & 0.9713     & 0.9785   & 0.9790   & \textbf{0.9752}   & \textbf{0.9805}       & \textbf{0.9729}    \\
Avg Runtime (s) & 7.64    & 6.28   & 5.90    & 5.85       & 7.71     & 32.56    & \textbf{5.64}     & \textbf{5.15}         & \textbf{3.30}      \\
\bottomrule
\end{tabular}
\end{adjustbox}
\end{table*}

\textbf{Benchmarking Performance and Basis Function Impact (RQ1).}
\label{sec:exp_mnist_bold}
This part evaluates SKANs with various single-parameter basis functions on MNIST, comparing them against other KANs and baselines for accuracy and efficiency. Experiments involved 5 runs with extensive learning rate (LR) searches (ranging from 0.00001 to 0.001, as KANs favor lower LRs), maintaining ~80k parameters, on NVIDIA 4060 Ti GPUs and i7-14700F CPUs. We tested SKAN variants (LSS, LArctan, LSin) against Spl-KAN, WavKAN, FastKAN, FourierKAN, and hybrid fKAN/rKAN models (which are less interpretable and MLP-dominant). Optimal F1 scores and average running time are in Table~\ref{tab:performance_comparison_single_row_adjusted} (also ploted in subplot c of \figurename~\ref{fig:overview}).

LArctan-SKAN achieved the highest F1 score, outperforming all KANs and hybrids, and ranked second in training speed, only behind LSin-SKAN, which were the fastest overall. LSin-SKAN also offered competitive accuracy with its superior speed. LArctan-SKAN was the top performer among activation-function-based SKANs, surpassing pure KANs in accuracy and speed. Key insights include:  LArctan-SKAN's F1 score (up to 6.51\% improvement over existing KANs, also nearly 6x faster than the slowest baseline) shows single-parameter basis potential; LSS-SKAN surpasses pure non-SKANs KAN variants in accuracy and speed; LSin is the most ideal for speed-critical tasks with speeds nearly 9x faster than the slowest baseline. SKANs can thus be chosen based on needs: LArctan for overall performance, LSS for stable activation-based accuracy, and LSin for speed.

\begin{figure*}[t]
    \centering
    \includegraphics[width=\linewidth]{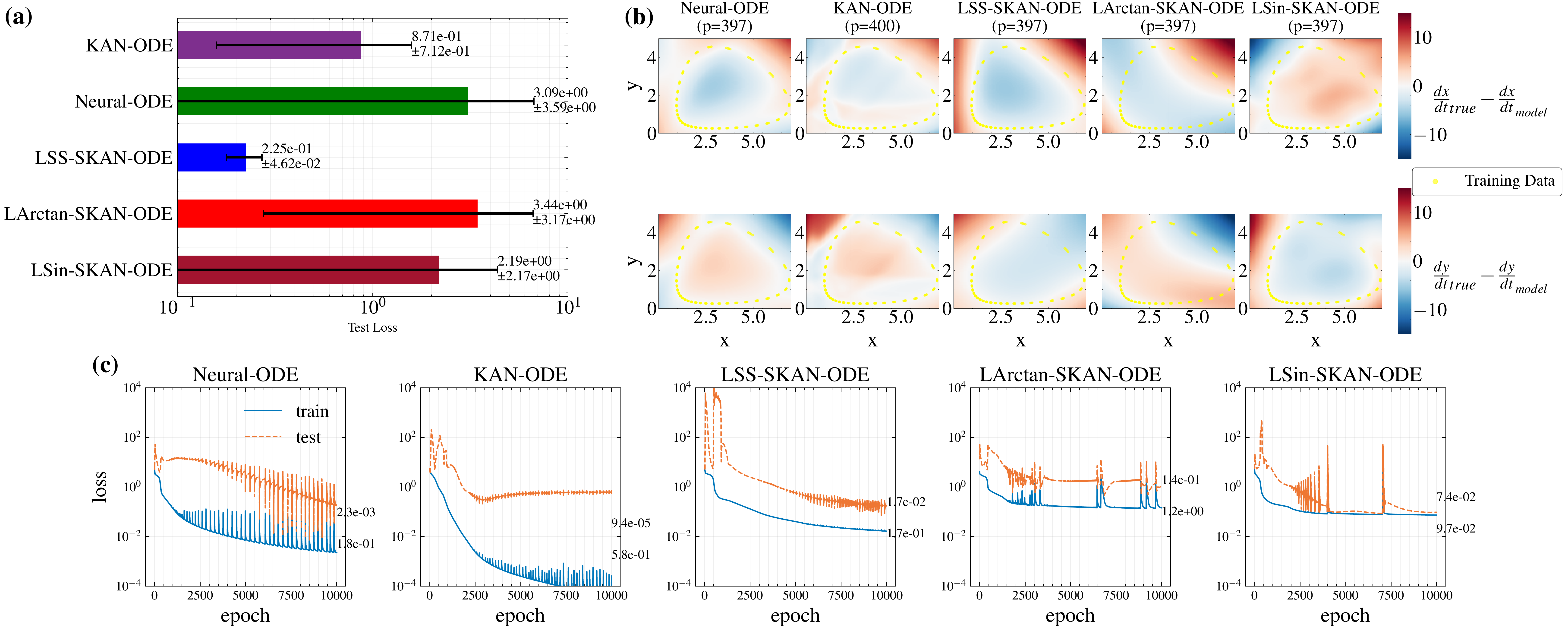}
    \caption{Performance comparison between SKAN-ODE and KAN-ODE.
    \textbf{(a)} Mean and standard deviation of test loss over 10 independent runs for each model.
    \textbf{(b)} Generalization performance visualization on the Lotka-Volterra system phase space from the median-performing run (ranked 5th in test loss), and \textbf{(c)} Training and test loss curves per epoch from the same median-performing run for each model.}
    \label{fig:ode_results} 
\end{figure*}

\begin{figure*}[t]
    \centering
    \includegraphics[width=\linewidth]{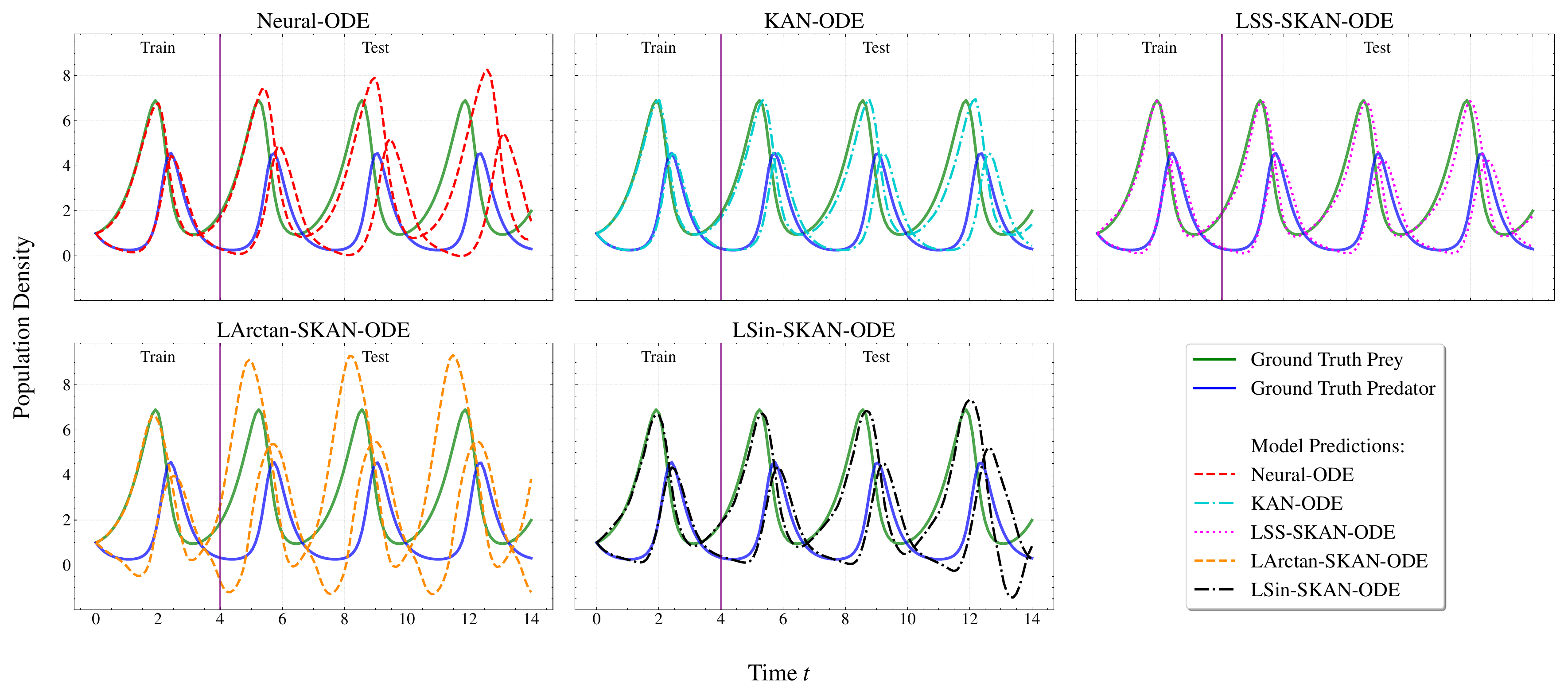}
    \caption{Lotka-Volterra Dynamics between Ground Truth and LSS-SKAN-ODE Predictions}
    \label{fig:ode_trad} 
\end{figure*}

\textbf{Domain Adaptability (RQ2).} To investigate the cross-domain adaptability of SKANs, this study systematically validates its performance across two critical scenarios: ordinary differential equation solving and medical image segmentation.

\textit{1) SKAN-ODE for Ordinary Differential Equation Solving.}
\label{sec:exp_ode_bold} We evaluated SKAN-ODE for solving ODEs. Adhering to the Lotka-Volterra setup (LR of 0.001, ~400 parameters) \cite{Koenig_2024}, we compared LSin-ODE, LArctan-ODE, and LSS-SKAN-ODE against Neural ODE and KAN-ODE. Each model underwent 10 runs of 10k epochs on CPUs. \figurename~\ref{fig:ode_results} shows (a) test loss statistics, (b) generalization visualizations, and (c) loss trajectories. Specifically, Subplot (b) visualizes model prediction errors against true dynamics, with training data overlaid.

As shown in subplot (a) of \figurename~\ref{fig:ode_results}, LSS-SKAN-ODE demonstrated superior mean test loss and stability, reducing loss by over 93.1\% versus Neural ODE and showing significantly lower standard deviation. Regarding generalization prediction, subplot (b) of \figurename~\ref{fig:ode_results} illustrates that LSS-SKAN-ODE and Neural ODE generalized best for $\frac{dy}{dt}$ predictions, while Neural ODE and KAN-ODE showed better extrapolation for $\frac{dx}{dt}$, with SKANs performing moderately and indicating room for improvement. Furthermore, subplot (c) of \figurename~\ref{fig:ode_results} reveals that KAN-ODE showed overfitting (consistent with \cite{Koenig_2024}), whereas LSS-SKAN-ODE achieved the lowest final test loss with a minimal train-test gap, signifying better accuracy and resistance to overfitting, and establishing it as the most stable and accurate SKAN-ODE variant. \figurename~\ref{fig:ode_trad} demonstrates that LSS-SKAN-ODE accurately predicts the Lotka-Volterra dynamics, closely matching the ground truth for both prey and predator populations.

\textit{2) SKAN for Medical Image Segmentation.}
\begin{table*}[t]
	\centering
	\caption{Comprehensive Performance Comparison for Medical Image Segmentation. All results are reported as mean $\pm$ std.dev. over 10 runs. Best performance in each category is highlighted in bold. Param. Change and Dice Improv. are calculated relative to the U-KAN baseline for each dataset.}
	\label{tab:full_segmentation_performance}
	\setlength{\tabcolsep}{4pt} 
	\begin{tabular}{cll r c c c c c} 
		\toprule
		Dataset & Model & Base Layer & \multicolumn{1}{c}{Parameters} & Param. Change (\%) & Dice / M Params & Loss$\downarrow$ & IoU$\uparrow$ & Dice Improv. (\%) \\
		\midrule
		\multirow{11}{*}{BUSI} & U-Net & MLP (Conv) & 6,356,689 & 0.00\% & 0.1156 & 0.470$\pm$0.023 & 0.596$\pm$0.014 & -0.14\% \\
		& U-KAN & Spl-KAN & 6,356,689 & \textit{Ref.} & 0.1158 & 0.469$\pm$0.024 & 0.595$\pm$0.016 & \textit{Ref.} \\
		& \textbf{U-LSS-SKAN} & LSS-SKAN & 6,273,671 & -1.31\% & \textbf{0.1195} & \textbf{0.440$\pm$0.014} & \textbf{0.613$\pm$0.013} & \textbf{+1.90\%} \\
		& U-LArctan-SKAN & LArctan-SKAN & 6,273,671 & -1.31\% & 0.1191 & 0.449$\pm$0.016 & 0.611$\pm$0.014 & +1.49\% \\
		& U-LSin-SKAN & LSin-SKAN & 6,273,671 & -1.31\% & 0.1173 & 0.476$\pm$0.079 & 0.595$\pm$0.043 & 0.00\% \\
		\cmidrule{2-9}
		& U-LSS-SKAN-275 & LSS-SKAN & 4,320,741 & -32.03\% & 0.1699 & 0.463$\pm$0.021 & 0.595$\pm$0.016 & -0.14\% \\
		& U-LArctan-SKAN-275 & LArctan-SKAN & 4,320,741 & -32.03\% & 0.1687 & 0.480$\pm$0.023 & 0.588$\pm$0.018 & -0.95\% \\
		& U-LSin-SKAN-275 & LSin-SKAN & 4,320,741 & -32.03\% & \textbf{0.1710} & 0.466$\pm$0.021 & 0.599$\pm$0.011 & \textbf{+0.41\%} \\
		\cmidrule{2-9}
		& U-LSS-SKAN-160 & LSS-SKAN & 2,764,561 & -56.51\% & 0.2637 & 0.491$\pm$0.025 & 0.586$\pm$0.014 & -0.95\% \\
		& U-LArctan-SKAN-160 & LArctan-SKAN & 2,764,561 & -56.51\% & \textbf{0.2673} & 0.475$\pm$0.022 & 0.600$\pm$0.013 & \textbf{+0.41\%} \\
		& U-LSin-SKAN-160 & LSin-SKAN & 2,764,561 & -56.51\% & 0.2669 & 0.476$\pm$0.019 & 0.599$\pm$0.012 & +0.27\% \\
		\midrule
		\multirow{5}{*}{CVC} & U-Net & MLP (Conv) & 6,356,689 & 0.00\% & 0.1430 & 0.207$\pm$0.016 & 0.838$\pm$0.030 & -0.33\% \\
		& U-KAN & Spl-KAN & 6,356,689 & \textit{Ref.} & 0.1435 & 0.203$\pm$0.017 & 0.842$\pm$0.026 & \textit{Ref.} \\
		& \textbf{U-LSS-SKAN} & LSS-SKAN & 6,273,671 & -1.31\% & \textbf{0.1460} & \textbf{0.192$\pm$0.013} & \textbf{0.848$\pm$0.026} & \textbf{+0.44\%} \\
		& U-LArctan-SKAN & LArctan-SKAN & 6,273,671 & -1.31\% & 0.1457 & 0.196$\pm$0.011 & 0.845$\pm$0.026 & +0.22\% \\
		& U-LSin-SKAN & LSin-SKAN & 6,273,671 & -1.31\% & \textbf{0.1460} & 0.194$\pm$0.012 & 0.847$\pm$0.027 & \textbf{+0.44\%} \\
		\midrule
		\multirow{5}{*}{GlaS} & U-Net & MLP (Conv) & 6,356,689 & 0.00\% & 0.1461 & 0.216$\pm$0.005 & 0.867$\pm$0.003 & +0.11\% \\
		& U-KAN & Spl-KAN & 6,356,689 & \textit{Ref.} & 0.1460 & 0.218$\pm$0.004 & 0.866$\pm$0.002 & \textit{Ref.} \\
		& \textbf{U-LSS-SKAN} & LSS-SKAN & 6,273,671 & -1.31\% & \textbf{0.1486} & \textbf{0.203$\pm$0.003} & \textbf{0.873$\pm$0.002} & \textbf{+0.43\%} \\
		& U-LArctan-SKAN & LArctan-SKAN & 6,273,671 & -1.31\% & 0.1484 & 0.206$\pm$0.004 & 0.871$\pm$0.002 & +0.32\% \\
		& U-LSin-SKAN & LSin-SKAN & 6,273,671 & -1.31\% & \textbf{0.1486} & 0.204$\pm$0.002 & \textbf{0.873$\pm$0.001} & \textbf{+0.43\%} \\
		\bottomrule
	\end{tabular}
\end{table*}
To assess the adaptability and parameter efficiency of SKANs in a high-stakes, real-world domain, we integrated the SKAN architecture into the U-Net framework, creating U-xx-SKAN variants. We conducted a rigorous evaluation on three public medical image segmentation benchmarks: BUSI \cite{ALDHABYANI2020104863}, CVC \cite{jha2019resunetadvancedarchitecturemedical}, and GlaS \cite{valanarasu2021medical}. Our evaluation was twofold: first, we compared full-sized U-SKANs-Net variants against U-Net and U-KAN baselines with comparable parameter counts. Second, to directly test the EKE hypothesis's implications for model efficiency, we engineered lightweight variants with substantially reduced parameter counts. All experiments adhered to the established protocol from \cite{Li_Liu_Li_Wang_Liu_Liu_Chen_Yuan_2025}, including 10 independent runs per configuration to ensure statistical validity. The comprehensive results are presented in Table~\ref{tab:full_segmentation_performance}.

The results unequivocally demonstrate the advantages of the SKAN design. At a comparable parameter scale, the U-LSS-SKAN variant consistently establishes itself as the top-performing model across all three datasets. On the BUSI dataset, it achieves a Dice score improvement of \textbf{+1.90\%} over the U-KAN baseline. This dominance, though more modest, is consistently maintained on the CVC and GlaS datasets, with improvements of \textbf{+0.44\%} and \textbf{+0.43\%}, respectively. This consistent superiority across diverse datasets underscores the robustness and general effectiveness of the LSS basis function for complex segmentation tasks.

Even more compelling is the performance of the parameter-reduced variants, which provides a powerful, real-world validation of the EKE hypothesis. On the BUSI dataset, the U-LArctan-SKAN-160 model, despite a staggering \textbf{56.51\% reduction in parameters}, not only remains competitive but surpasses the full-sized U-Net and U-KAN baselines, achieving a Dice improvement of \textbf{+0.41\%} over U-KAN. This remarkable achievement is further quantified by the parameter efficiency metric, `Dice / M Params', where the U-LArctan-SKAN-160 scores an impressive 0.2673—more than double the efficiency of the baselines. This demonstrates that by embracing simpler basis functions, it is possible to construct significantly lighter models that are not only more efficient but also more effective. This finding is of paramount importance in domains like medical imaging, where the deployment of accurate, low-resource models is a critical objective.

\textbf{More Discussion on the EKE Hypothesis (RQ3).}
\label{sec:exp_eke_validation_bold}
The EKE hypothesis is strongly supported by our experimental findings. On MNIST, SKANs like LArctan-SKAN, using single-parameter basis, achieved superior F1 scores and competitive speeds versus KANs with more complex basis (e.g., Spl-KAN). This aligns with EKE's proposition of effective parameter use through simpler basis, further supported by SKANs' computational efficiency.

This validation extends to specialized domains. In ODE solving, LSS-SKAN-ODE outperformed KAN-ODE (using complex splines) and Neural ODE in test loss and stability, often with design choices favored by EKE. In medical image segmentation, LSS-SKAN-UNet surpassed U-Net and U-KAN. Crucially, reduced-parameter SKANs-UNet (e.g., LArctan-SKAN-UNet-160) achieved comparable or better F1 scores than full-parameter baselines using substantially fewer parameters (56.51\% less), directly showing EKE's predicted parameter efficiency.

Collectively, results from diverse tasks show that leveraging ultra-lightweight basis functions to prioritize network architecture is a potent strategy. SKANs' success in performance, speed, and efficiency validates EKE as a key guideline for KAN development. 

\section{Conclusion and Future Work}
\textbf{Conclusion.}
This paper introduces the Uni-KAN framework, a formal, unified abstraction for the burgeoning field of Kolmogorov-Arnold Networks. Building on this framework, we propose the EKE hypothesis, which challenges the prevailing focus on basis function complexity and advocates instead for prioritizing architectural scaling. This hypothesis leads to the development of SKANs, a novel family of ultra-lightweight networks that utilize single-parameter basis functions. The superior performance of SKANs is validated through extensive experiments in multiple domains, including handwritten digit recognition, differential equation solving, and medical image segmentation. In summary, the Uni-KAN framework, the EKE hypothesis, and the SKAN architecture collectively establish a new paradigm for KAN research, offering a unified theoretical foundation, a clear design philosophy, and a practical path toward more efficient and powerful models.

\textbf{Future Work.}
Although this study briefly discusses the interpretability of SKAN through visualization, a more comprehensive analysis is warranted. The decoupling and isolation of nonlinear units in pure KAN variants inherently enhances model interpretability, and SKAN preserves this property. Future research should further explore and quantify the interpretability advantages of SKAN.
Meanwhile, the generalization performance of SKAN-ODE in solving differential equations leaves room for improvement. Enhancing the extrapolation capabilities of SKAN-ODE beyond the training domain is an important direction for future work.
Additionally, discovering additional high-performance single-parameter functions remains an open challenge. Continued exploration in this area may yield basis functions that further optimize the trade-off between network scale and basis function complexity. 

\bibliography{ref}
\bibliographystyle{unsrt}

\end{document}